# Constrained Extreme Learning Machines: A Study on Classification Cases

Wentao Zhu, Jun Miao, and Laiyun Qing

*Abstract*—Extreme learning machine (ELM) is an extremely fast learning method and has a powerful performance for pattern recognition tasks proven by enormous researches and engineers. However, its good generalization ability is built on large numbers of hidden neurons, which is not beneficial to real time response in the test process. In this paper, we proposed new ways, named "constrained extreme learning machines" (CELMs), to randomly select hidden neurons based on sample distribution. Compared to completely random selection of hidden nodes in ELM, the CELMs randomly select hidden nodes from the constrained vector space containing some basic combinations of original sample vectors. The experimental results show that the CELMs have better generalization ability than traditional ELM, SVM and some other related methods. Additionally, the CELMs have a similar fast learning speed as ELM.

## I. INTRODUCTION

MANY neural network architectures have been developed over the past several decades. The feedforward neural networks are especially the most popular ones studied all the times. It has been proven that the learning capacity of a multilayer feedforward neural network with non-polynomial activation functions can approximate any continuous function [1]. Single hidden layer feedforward neural networks (SLFNs) was studied and applied extensively by researches because of their model simplicity and relatively high learning and responding speed. The learning capacity of SLFNs is not inferior to that of multilayer feedforward neural networks, as proved in [2, 3]. There are mainly three different ways to train the SLFNs:

1. Gradient based learning methods. The most famous gradient based learning method is back-propagation algorithm [4]. However, these methods may face quite slow learning speed and stack into local minimal. Although many assistant methods were proposed to solve such problems, such as Levenberg-Marquardt (LM) method [5], dynamically network construction [6], evolutionary algorithms [7] and generic optimization [8], the enhanced methods require heavy computation or cannot obtain a global optimal solution.
2. Optimization based learning methods. One of the most popular optimization based SLFNs is Support Vector Machine (SVM) [9]. The objective function of SVM is to optimize the weights for maximum margin corresponding to structural risk minimization. The solution of SVM can be obtained by convex optimization methods in the dual problem space and is the global optimal solution. SVM is a very popular method attracting many researchers [10].
3. Least Mean Square (LMS) based methods, such as Radial Basis Function network [11] and No-Prop network [12] based on LMS algorithm [13]. These methods have quite fast learning speed due to the essence of "No-Prop" and fast learning speed of LMS.

In recent years, Huang et al. [14] proposed a novel extremely fast learning model of SLFNs, called Extreme Learning Machine (ELM). One of its salient essences that the weights from the input layer to the hidden layer can be randomly generated was firstly shown by Tamura et al. [2]. Huang et al. [14] further completely proved the random feature mapping theory rigorously.

After the random nonlinear feature mapping in the hidden layer, the rest of ELM can be considered as a linear system [15]. Therefore, ELM has a closed form of solution due to the simple network structure and random hidden layer weights. The essence of the linear system used by ELM is to minimize the training error and the norm of connection weights from the hidden layer to the output layer at the same time [16]. Hence ELM has a good generalization performance according to the feedforward neural network theory [12, 17]. As a consequence, ELM has some desirable features, such as that hidden layer parameters need not be tuned, fast learning speed and good generalization performance. Additionally, ELM has a unified framework for classification, regression, semi-supervised, supervised and unsupervised tasks [16, 18]. These advantages lead to the popularity of ELM both for researchers and engineers [19, 20, 21, 22, 54].

However, the random selection of hidden layer parameters makes quite inefficient use of hidden nodes [23]. ELM usually has to randomly generate a great number of hidden nodes to achieve desirable performance. This leads to time consuming in test process, which is not helpful in real applications. Large numbers of hidden nodes also easily make the trained model

This work was supported in part by Natural Science Foundation of China (Nos. 61175115 and 61272320) and President Fund of Graduate University of Chinese Academy of Sciences (No. Y35101CY00).

W. Zhu and J. Miao are with the Key Lab of Intelligent Information Processing of Chinese Academy of Sciences (CAS), Institute of Computing Technology, CAS, Beijing 100190, China (e-mail: wentaozhu1991@gmail.com).

L. Qing is with School of Computer and Control Engineering, University of Chinese Academy of Sciences, Beijing 100049, China.

stack into over fitting. There are mainly three ways to solve the problem for a more compact model:

1. Use online incremental learning methods to add hidden layer nodes dynamically [23, 24]. These methods randomly generate parts or all of the hidden layer nodes, and then select the candidate hidden nodes one by one or chunk by chunk with fixed or varying chunk size. Whether the hidden layer node is added or not is usually depending on the objective function of the output layer.
2. Use pruning methods to select the candidate hidden layer nodes [25, 26]. These methods start with a large neural network using the traditional ELM, and then apply some metrics, such as statistical criteria and multi-response sparse regression, to rank these hidden nodes. Finally, eliminate those low relevance hidden nodes to form a more compact neural network structure.
3. Use gradient based methods to update the weights from the input layer to the hidden layer in ELM [27]. After randomly initialize the weights from the input layer to the hidden layer and use a close-form least square solution to calculate the weights from the hidden layer to the output layer, these methods use the gradient descending method to update the weights from the input layer to the hidden layer in ELM iteratively.

The above methods can overcome the drawbacks of the traditional ELM to some degree. However, they do not solve the problem directly from the essence of hidden nodes. Besides, these methods are somewhat time-consuming.

The essence of hidden layer functions is to map the data into a feature space, where the output layer can use a linear classifier to separate the data perfectly. Therefore, the hidden layer should extract discriminative features or some other data representations for classification tasks.

LDA [28] is probably the most commonly used method to extract discriminative features. However, traditional LDA has some drawbacks, such as that the number of feature-mapped dimensions is less than the number of classes, "Small Sample Size" (SSS) problem and Gaussian distribution assumption of equal covariance and different means. Su et al. [29] proposed a projection pursuit based LDA method to overcome these problems. The method showed that the difference vectors of between-class samples have a strong discriminative property for classification tasks, but this method is rather complex with many embedded trivial tricks.

In this work, to balance the high discriminative feature learning and the fast training speed of the ELM, we propose a novel model, called Constrained difference Extreme Learning Machine (CDELM), which utilizes a random subset of difference vectors of between-class samples to replace the completely random connection weights from the input layer to the hidden layer in ELM. More generally, the linear combination of sample vectors, such as sum vectors of within-class samples, sample vectors of all classes, sum vectors of randomly selected sample vectors and the mixed vectors including difference vectors of between-class samples and sum vectors of within-class samples, as connection weights from the input layer to the hidden layer are validated. We proposed Constrained Sum Extreme Learning Machine (CSELM), Sample Extreme Learning Machine (SELM), Random Sum Extreme Learning Machine (RSELM) and Constrained Mixed Extreme Learning Machine (CMELM) based on these data-driven hidden layer features mapping ways. Experimental results show that, CELMs has better generalization ability than ELM related methods, SVM related methods and the BP neural network, whilst retaining the fast learning characteristics of ELM. We also compared the CELM algorithms with SVM and ELM related algorithms [30, 31] on CIFAR-10 data set [32]. The results show that the CELM algorithms outperform these methods.

The remaining part of the paper is organized as follows: in section II, we review the traditional ELM algorithm, and then propose the CELMs in section III. Experiments are presented in section IV. Conclusion and discussion are given in section V.

## II. REVIEW OF EXTREME LEARNING MACHINE

ELM is a type of SLFNs. The hidden layer parameters, i.e., the connection weights from the input layer to the hidden nodes, are randomly generated in ELM. The output layer is a linear system, where the connection weights from the hidden layer to the output layer are learned by computing the Moore-Penrose generalized inverse [14]. The ELM network has an extreme high learning speed due to the simple network structure and its closed form solution. Additionally, the randomness makes ELM not necessarily tune these hidden layer parameters iteratively.

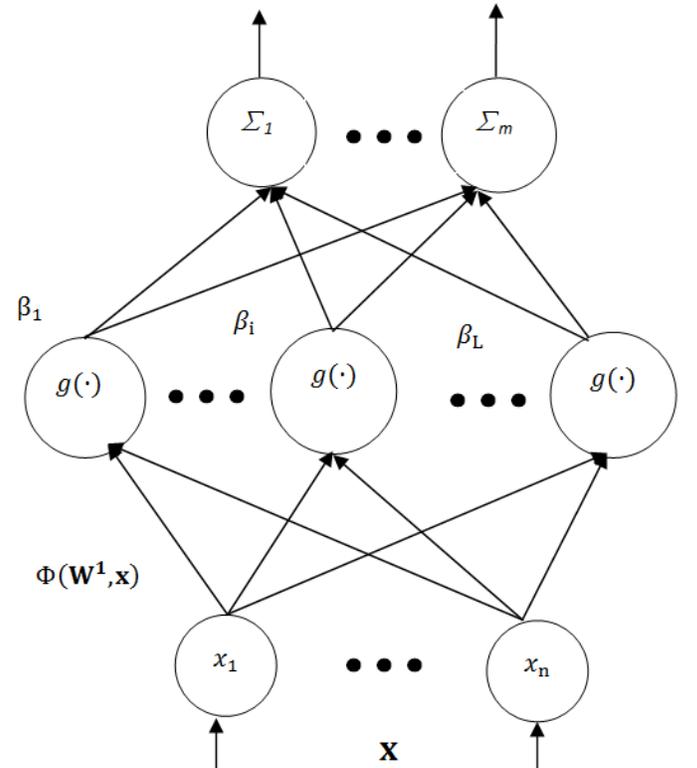

Fig. 1. Illustration of the structure of ELM neural network.

Given the training samples and class labels

$\aleph = \{(\mathbf{x}_i, \mathbf{t}_i) | \mathbf{x}_i \in \mathbf{R}^n, \mathbf{t}_i \in \mathbf{R}^m, i=1,\ldots,N\}$, the number of hidden nodes $L$ and activation function $G(\mathbf{a},b,\mathbf{x})$, where $\mathbf{x} \in \mathbf{R}^n$ is the input vector, $\mathbf{a} \in \mathbf{R}^n$ is the associated connection weight vector and $b \in \mathbf{R}$ is the bias, the algorithm of ELM network can be summarized as the following three steps:

Step 1: Assign the parameters $\{(\mathbf{a}_j, b_j) | j=1,\ldots,L\}$ of hidden nodes with randomly generated values.

Step 2: Calculate the hidden layer output matrix H for all the training samples:

$$\mathbf{H} = \begin{bmatrix} \mathbf{h}(\mathbf{x}_1) \\ \vdots \\ \mathbf{h}(\mathbf{x}_N) \end{bmatrix} = \begin{bmatrix} G(\mathbf{a}_1,b_1,\mathbf{x}_1) & \cdots & G(\mathbf{a}_L,b_L,\mathbf{x}_1) \\ \vdots & \cdots & \vdots \\ G(\mathbf{a}_1,b_1,\mathbf{x}_N) & \cdots & G(\mathbf{a}_L,b_L,\mathbf{x}_N) \end{bmatrix}_{N \times L}$$

, where $G(\mathbf{a}_i, b_i, \mathbf{x}_j)$ is the activation function of $i$th hidden node for $j$th sample.

Step 3: Calculate the hidden layer's output connection weights $\beta$ by solving the least squares problem:

$$\beta = \mathbf{H}^\dagger \mathbf{T}$$

, where $\mathbf{H}^\dagger$ is the generalized inverse matrix of the matrix $\mathbf{H}$, and $\mathbf{T} = \begin{bmatrix} \mathbf{t}_1^T \\ \vdots \\ \mathbf{t}_N^T \end{bmatrix}_{N \times m}$.

However, the condition number of the random projected matrix $\mathbf{H}$ may be very large and the above traditional ELM model may encounter ill-posed problems [33]. In practice, regularized term with hidden layer's output connection weights $\beta$ is added into the optimization objective to avoid the problem [34, 35, 36]. The solution of regularized ELM can be obtained as

$$\beta = \mathbf{H}^T (\frac{\mathbf{I}}{\lambda} + \mathbf{H}\mathbf{H}^T)^{-1} \mathbf{T}$$

$$\text{or } \beta = (\frac{\mathbf{I}}{\lambda} + \mathbf{H}^T\mathbf{H})^{-1} \mathbf{H}^T \mathbf{T}$$

, where $\mathbf{I}$ is the identity matrix and $\lambda$ is the regularization factor which can be obtained by cross validation in the training process.

As analyzed in theory and further verified by the simulation results in [34], ELM for classification tends to achieve better generalization performance than traditional SVM. ELM can also overcome the local minimal problem that BP neural nets faced, due to its convex model structure. The learning speed of ELM is extremely fast at the same time.

## III. CONSTRAINED EXTREME LEARNING MACHINES

In this section, we will introduce the CELMs with the idea of using simple linear combination of sample vectors to generate hidden nodes in the traditional ELM network structure.

### A. Constrained Difference Extreme Learning Machine

The Constrained Difference Extreme Learning Machine (SCELM) attempts to extract discriminative features in the hidden layer. The completely random parameters in the hidden layer of ELM do not always represent discriminative features. Such unconstrained random parameters may make ELM has to generate a great number of hidden nodes to meet desirable generalization performances. More hidden nodes mean more processing time and more easily over fitting. These problems in ELM should be solved.

Although the method [29] is rather complex with many embedded trivial tricks, it shows that the difference vectors of between-class samples are effective to classification tasks. Considering the simplicity and the extreme high learning speed of the ELM, we extend the ELM model to Constrained Difference Extreme Learning Machine (CDELM) by constraining the weight vector parameters $\{\mathbf{a}_j | j=1,\ldots,L\}$ of ELM to be randomly drawn from a closed set of difference vectors of between-class samples instead of from the open set of arbitrary vectors to tackle the problem of generation of discriminative hidden nodes. We use a simple case to illustrate the idea of difference vectors of between-class samples.

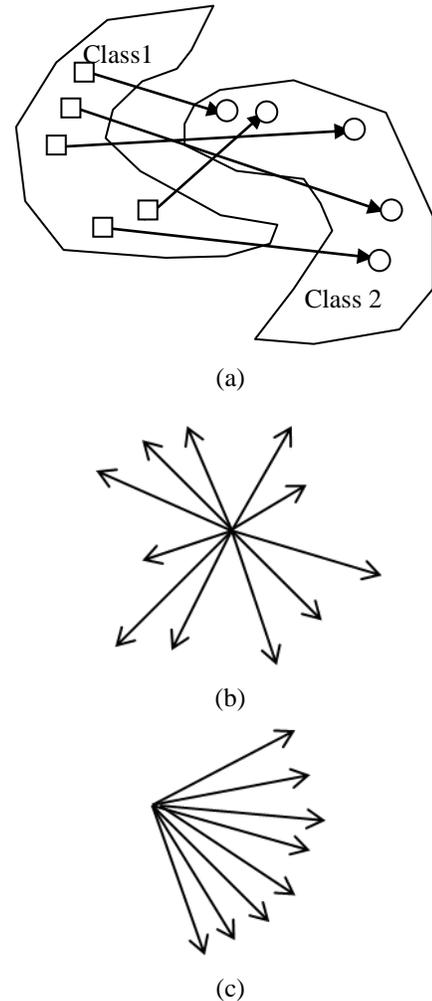

Fig. 2. Illustration of difference vectors of between-class samples. How to generate the difference vectors is illustrated in (a). The completely random connection weight vectors from the input layer to the hidden layer of ELM are illustrated in (b). The constrained random weight vectors of CELM are illustrated in (c).

The essence of the weight vectors from the input layer to the neurons in the hidden layer is to map the original samples into a discriminative feature space spanned by these vectors, where the samples can be classified. The weight vectors are helpful for classification if the directions of the weight vectors are from class 1 to class 2 or reversely, as illustrated in Fig. 1(a). The blocks in the figure represent the samples in class 1, and the circles represent the samples in class 2. As a comparison, the weight vectors from the input layer to the hidden nodes in ELM are completely random without constraints, as illustrated in Fig. 1(b). It can be inferred the weight vectors which do not follow the direction from class 1 to class 2 are less discriminative for classification tasks. This is the reason why not all the hidden nodes in ELM are efficient or discriminative for classification tasks.

We randomly generate the weight vectors from the input layer to the hidden layer with the differences of between-class samples as illustrated in Fig. 1(a). The difference vectors of between-class samples can map the samples to a higher discriminative feature space than ELM. The weight vectors from the input layer to the hidden layer in CDELM are illustrated in Fig. 1(c). The directions of these weight vectors are close to the direction from class 1 to class 2, which are more discriminative for the classification tasks intuitively.

We normalize the difference vectors as the weights from the input layer to the hidden layer. The reason why normalize these weights and how to normalize will be introduced in the following discussion. In our originally Constrained Difference Extreme Learning Machine [37], we deleted the difference vectors of small norms and too relevant difference vectors. These pre-processing operations are somewhat time-consuming especially when the number of hidden nodes is large. Although the two processes can improve the performance [38], the improvement is very little in our experiments.

In CDELM, the prior information of samples' class distribution is utilized to generate the weights from the input layer to the hidden layer. The aim is to split different classes' samples into different areas in the feature space. The ideal case is that, for example, class 1 is mapped into negative semi axis and class 2 is mapped into positive semi axis in the feature space. Hence the bias must be set as the middle point of the two selected samples from the geometric sense intuitively. As a result, the biases to the hidden neurons can be determined by assuming that the samples from one class are mapped to -1 and the samples from another class are mapped 1 respectively. Denote $\mathbf{x}_{c1}$ and $\mathbf{x}_{c2}$ as the samples drawn from two classes. Then the weight vector $\mathbf{w}$ from the input layer to one hidden neuron can be generated with $\alpha(\mathbf{x}_{c2} - \mathbf{x}_{c1})$, where $\alpha$ is the normalized factor. The original data $\mathbf{x}$ is transformed to $\mathbf{x}^T\mathbf{w} + b = \alpha \mathbf{x}^T(\mathbf{x}_{c2} - \mathbf{x}_{c1}) + b$ by feature mapping, where $b$ is the bias with respect to the weight vector $\mathbf{w}$ in ELM model. The assumption that $\mathbf{x}_{c1}$ and $\mathbf{x}_{c2}$ are mapped to -1 and 1 can be written as

$$\alpha \mathbf{x}_{c1}^T (\mathbf{x}_{c2} - \mathbf{x}_{c1}) + b = -1, \text{ and}$$
$$\alpha \mathbf{x}_{c2}^T (\mathbf{x}_{c2} - \mathbf{x}_{c1}) + b = 1.$$

We can obtain that the normalization factor $\alpha = \dfrac{2}{\|\mathbf{x}_{c2} - \mathbf{x}_{c1}\|_{L_2}^2}$ and the corresponding bias $b = \dfrac{(\mathbf{x}_{c1} + \mathbf{x}_{c2})^T(\mathbf{x}_{c1} - \mathbf{x}_{c2})}{\|\mathbf{x}_{c2} - \mathbf{x}_{c1}\|_{L_2}^2}$ by solving the above two equation constraints.

The commonly used activation function for hidden neurons is sigmoid function $f(x) = \dfrac{1}{1+e^{-x}}$. The output layer in CDELM is a simple linear system as same as that of ELM.

From the above discussion, the training algorithm for CDELM can be concluded in the Algorithm 1. The essence of CDELM is to constrain the hidden neuron's input connection weights to be consistent with the directions from one class to another class. So the random weights are constrained to be chosen from the set that is composed of the difference vectors of between-class samples.

---

**Algorithm 1:** Training of the Constrained Difference Extreme Learning Machine (CDELM)

**Input:** the training samples $\aleph = \{(\mathbf{x}_i, \mathbf{t}_i) | \mathbf{x}_i \in \mathbf{R}^n, \mathbf{t}_i \in \mathbf{R}^m, i = 1, \ldots, N\}$, the hidden node number $L$ and the activation function $G(\mathbf{w}, b, \mathbf{x})$.

**Output:** the model parameters of CELM, i.e., the weight matrix $\mathbf{W}_{n \times L}$ and the bias vector $\mathbf{b}_{1 \times L}$ from the input layer to the hidden layer, the weight matrix $\beta_{L \times m}$ from the hidden layer to the output layer.

1) While the number of chosen difference vectors is less than $L$

   a) Randomly draw training samples $\mathbf{x}_{c1}$ and $\mathbf{x}_{c2}$ from any two different classes respectively and generate the difference vector $\mathbf{x}_{c2} - \mathbf{x}_{c1}$;

   b) Normalize the difference vector by $\mathbf{w} = \dfrac{2(\mathbf{x}_{c2} - \mathbf{x}_{c1})}{\|\mathbf{x}_{c2} - \mathbf{x}_{c1}\|_{L_2}^2}$, and calculate the corresponding bias $b = \dfrac{(\mathbf{x}_{c1} + \mathbf{x}_{c2})^T(\mathbf{x}_{c1} - \mathbf{x}_{c2})}{\|\mathbf{x}_{c2} - \mathbf{x}_{c1}\|_{L_2}^2}$.

   c) Use the vector $\mathbf{w}$ and bias $b$ to construct the weight matrix $\mathbf{W}_{n \times L}$ and bias vector $\mathbf{b}_{1 \times L}$.

2) Calculate the hidden layer output matrix $\mathbf{H}$ as

$$\mathbf{H} = \begin{bmatrix} \mathbf{h}(\mathbf{x}_1) \\ \vdots \\ \mathbf{h}(\mathbf{x}_N) \end{bmatrix} = \begin{bmatrix} G(\mathbf{a}_1, b_1, \mathbf{x}_1) & \cdots & G(\mathbf{a}_L, b_L, \mathbf{x}_1) \\ \vdots & \cdots & \vdots \\ G(\mathbf{a}_1, b_1, \mathbf{x}_N) & \cdots & G(\mathbf{a}_L, b_L, \mathbf{x}_N) \end{bmatrix}_{N \times L}.$$

3) Calculate the hidden layer's output weight matrix $\beta_{L \times m}$ by solving the least squares problem:

$$\beta = \mathbf{H}^\dagger \mathbf{T}$$

, where $\mathbf{H}^\dagger$ is the generalized inverse matrix and $\mathbf{T} = \begin{bmatrix} \mathbf{t}_1^T \\ \vdots \\ \mathbf{t}_N^T \end{bmatrix}_{N \times m}$.

### B. Sample Extreme Learning Machine

The Sample Extreme Learning Machine (SELM) utilizes sample vectors that are randomly drawn from the training set to construct the weights from the input layer to the hidden layer. The SELM firstly randomly selects a group of sample vectors, then normalizes these vectors as $\frac{\mathbf{x}_i}{\|\mathbf{x}_i\|_{L_2}^2}$, where $\mathbf{x}_i$ is the chosen sample vector. After the normalization, the sample itself can be transformed of norm 1 to serve the linear classification in the output layer. The normalized sample vectors are assigned as the weights from the input layer to the hidden layer. The biases used in SELM are randomly generated from uniform distribution as same as that in ELM.

The SELM model is a little like kernel based methods. If the activation function is sigmoid function, we can formula the $i$th hidden node in SELM as

$$G(\mathbf{a}_i, b_i, \mathbf{x}) = G(\mathbf{x}_i, b_i, \mathbf{x}) = \frac{1}{1+\exp[-(\mathbf{x}_i \mathbf{x} + b_i)]}$$

, where $\mathbf{x}$ is one input vector, $\mathbf{x}_i$ is the $i$th hidden node weight, which is selected randomly from sample vectors. Actually, $K(\mathbf{x}_i, \mathbf{x}) = \mathbf{x}_i \mathbf{x} + b_i$ is a polynomial kernel function, which can tackle some linearly inseparable cases, such as data sets of quadratic curve. The difference between the kernel used here and SVM kernel is that, the samples in the SVM kernel function is the support vectors. The difference between kernel ELM [16] and our SELM is that ELM kernel uses all the training samples. The sigmoidal activation function used is to stretch the kernel mapped data and helps the linear classification in the output layer.

From the above discussion, we can write the SELM algorithm as Algorithm 2. The essence of SELM is to constrain the hidden neuron's input connection weights to be consistent with the directions of sample vectors. So the original random weights are constrained to be chosen from the training set.

**Algorithm 2:** Training of the Sample Extreme Learning Machine (SELM)

**Input:** the training samples $\aleph = \{(\mathbf{x}_i, \mathbf{t}_i) | \mathbf{x}_i \in \mathbf{R}^n, \mathbf{t}_i \in \mathbf{R}^m, i=1,\ldots,N\}$, the hidden node number $L$ and the activation function $G(\mathbf{w}, b, \mathbf{x})$.

**Output:** the model parameters of SELM, i.e., the weight matrix $\mathbf{W}_{n \times L}$ and the bias vector $\mathbf{b}_{1 \times L}$ from the input layer to the hidden layer, the weight matrix $\beta_{L \times m}$ from the hidden layer to the output layer.

1) While the number of chosen sample vectors is less than $L$
  a) Randomly draw samples $\mathbf{x}_i$ from the training data;
  b) Normalize the sample vector by $\mathbf{w} = \frac{\mathbf{x}_i}{\|\mathbf{x}_i\|_{L_2}^2}$, and draw the corresponding bias $b$ randomly from [0,1] uniform distribution.
  c) Use the vector $\mathbf{w}$ and bias $b$ to construct the weight matrix $\mathbf{W}_{n \times L}$ and bias vector $\mathbf{b}_{1 \times L}$.

2) Calculate the hidden layer output matrix $\mathbf{H}$ as

$$\mathbf{H} = \begin{bmatrix} \mathbf{h}(\mathbf{x}_1) \\ \vdots \\ \mathbf{h}(\mathbf{x}_N) \end{bmatrix} = \begin{bmatrix} G(\mathbf{a}_1, b_1, \mathbf{x}_1) & \cdots & G(\mathbf{a}_L, b_L, \mathbf{x}_1) \\ \vdots & \cdots & \vdots \\ G(\mathbf{a}_1, b_1, \mathbf{x}_N) & \cdots & G(\mathbf{a}_L, b_L, \mathbf{x}_N) \end{bmatrix}_{N \times L}$$

3) Calculate the hidden layer's output weight matrix $\beta_{L \times m}$ by solving the least squares problem:

$$\beta = \mathbf{H}^\dagger \mathbf{T}$$

, where $\mathbf{H}^\dagger$ is the generalized inverse matrix and $\mathbf{T} = \begin{bmatrix} \mathbf{t}_1^T \\ \vdots \\ \mathbf{t}_N^T \end{bmatrix}_{N \times m}$.

### C. Constrained Sum Extreme Learning Machine

The Constrained Sum Extreme Learning Machine (CSELM) utilizes sum vectors of random chosen within-class sample vectors to construct the weights from the input layer to hidden layer. The CSELM firstly randomly selects any two within-class sample vectors $\mathbf{x}_c'$ and $\mathbf{x}_c''$, calculate the sum of the two vectors $\mathbf{x}_c' + \mathbf{x}_c''$, then normalizes the sum vector as $\frac{\mathbf{x}_c' + \mathbf{x}_c''}{\|\mathbf{x}_c' + \mathbf{x}_c''\|_{L_2}^2}$. The normalized sum sample vectors are assigned as the weights from the input layer to the hidden layer. The biases used in CSELM are also randomly generated from the uniform distribution as same as that in ELM. The constrained sum vectors used here were firstly inspired by the difference vectors of between-class samples. The constrained sum vectors also can be considered as some derivative samples, which can somewhat weaken the affection of noise samples in CSELM.

From the above discussion, we can design the SELM algorithm as Algorithm 3. The essence of CSELM is to constrain the hidden neuron's input connection weights to be consistent with the directions of derivative robust sample vectors. So the random weights are constrained to be chosen from the set that is composed of the sum vectors of within-class sample vectors.

**Algorithm 3:** Training of the Constrained Sum Extreme Learning Machine (CSELM)

**Input:** the training samples $\aleph = \{(\mathbf{x}_i, \mathbf{t}_i) | \mathbf{x}_i \in \mathbf{R}^n, \mathbf{t}_i \in \mathbf{R}^m, i=1,\ldots,N\}$, the hidden node number $L$ and the activation function $G(\mathbf{w}, b, \mathbf{x})$.

**Output:** the model parameters of CSELM, i.e., the weight matrix $\mathbf{W}_{n \times L}$ and the bias vector $\mathbf{b}_{1 \times L}$ from the input layer to the hidden layer, the weight matrix $\beta_{L \times m}$ from the hidden layer to the output layer.

1) While the number of chosen constrained sum vectors is less than $L$

a) Randomly draw training samples $\mathbf{x}_c'$ and $\mathbf{x}_c''$ from the same class respectively and generate the sum vector $\mathbf{x}_c' + \mathbf{x}_c''$;

b) Normalize the sum vector by $\mathbf{w} = \dfrac{\mathbf{x}_c' + \mathbf{x}_c''}{\left\| \mathbf{x}_c' + \mathbf{x}_c'' \right\|_{L_2}^2}$, and draw the corresponding bias $b$ randomly from [0,1] uniform distribution.

c) Use the vector $\mathbf{w}$ and bias $b$ to construct the weight matrix $\mathbf{W}_{n \times L}$ and bias vector $\mathbf{b}_{1 \times L}$.

2) Calculate the hidden layer output matrix $\mathbf{H}$ as

$$\mathbf{H} = \begin{bmatrix} \mathbf{h}(\mathbf{x}_1) \\ \vdots \\ \mathbf{h}(\mathbf{x}_N) \end{bmatrix} = \begin{bmatrix} G(\mathbf{a}_1, b_1, \mathbf{x}_1) & \cdots & G(\mathbf{a}_L, b_L, \mathbf{x}_1) \\ \vdots & \cdots & \vdots \\ G(\mathbf{a}_1, b_1, \mathbf{x}_N) & \cdots & G(\mathbf{a}_L, b_L, \mathbf{x}_N) \end{bmatrix}_{N \times L}.$$

3) Calculate the hidden layer's output weight matrix $\beta_{L \times m}$ by solving the least squares problem:

$$\beta = \mathbf{H}^\dagger \mathbf{T}$$

, where $\mathbf{H}^\dagger$ is the generalized inverse matrix and $\mathbf{T} = \begin{bmatrix} \mathbf{t}_1^T \\ \vdots \\ \mathbf{t}_N^T \end{bmatrix}_{N \times m}$.

### D. Random Sum Extreme Learning Machine

The Random Sum Extreme Learning Machine (RSELM) utilizes sum vectors of random sample vectors regardless of classes to construct the weights from the input layer to the hidden layer. The RSELM firstly randomly selects any two sample vectors $\mathbf{x}'$ and $\mathbf{x}''$, calculate the sum of the two vectors $\mathbf{x}' + \mathbf{x}''$, then normalizes the sum vector as $\dfrac{\mathbf{x}' + \mathbf{x}''}{\left\| \mathbf{x}' + \mathbf{x}'' \right\|_{L_2}^2}$. The normalized sum vectors are assigned as the weights from input layer to the hidden layer. The biases used in RSELM are also randomly generated from the uniform distribution as same as that in ELM. The sum vectors of random samples used here is to accelerate the speed of hidden layer weights generation.

From the above discussion, we can design the RSELM algorithm as Algorithm 4. The essence of RSELM is to constrain the hidden neuron's input connection weights to be consistent with the sum vectors of random samples.

---

**Algorithm 4:** Training of the Random Sample Extreme Learning Machine (RSELM)

**Input:** the training samples $\aleph = \{(\mathbf{x}_i, \mathbf{t}_i) | \mathbf{x}_i \in \mathbf{R}^n, \mathbf{t}_i \in \mathbf{R}^m, i = 1, \ldots, N\}$, the hidden node number $L$ and the activation function $G(\mathbf{w}, b, \mathbf{x})$.

**Output:** the model parameters of CELM, i.e., the weight matrix $\mathbf{W}_{n \times L}$ and the bias vector $\mathbf{b}_{1 \times L}$ from the input layer to the hidden layer, the weight matrix $\beta_{L \times m}$ from the hidden layer to the output layer.

1) While the number of chosen random sum vectors is less than $L$

a) Randomly draw training samples $\mathbf{x}'$ and $\mathbf{x}''$ from data samples;

b) Normalize the difference vector by $\mathbf{w} = \dfrac{\mathbf{x}' + \mathbf{x}''}{\left\| \mathbf{x}' + \mathbf{x}'' \right\|_{L_2}^2}$, and draw the corresponding bias $b$ randomly from [0,1] uniform distribution.

c) Use the vector $\mathbf{w}$ and bias $b$ to construct the weight matrix $\mathbf{W}_{n \times L}$ and bias vector $\mathbf{b}_{1 \times L}$.

2) Calculate the hidden layer output matrix $\mathbf{H}$ as

$$\mathbf{H} = \begin{bmatrix} \mathbf{h}(\mathbf{x}_1) \\ \vdots \\ \mathbf{h}(\mathbf{x}_N) \end{bmatrix} = \begin{bmatrix} G(\mathbf{a}_1, b_1, \mathbf{x}_1) & \cdots & G(\mathbf{a}_L, b_L, \mathbf{x}_1) \\ \vdots & \cdots & \vdots \\ G(\mathbf{a}_1, b_1, \mathbf{x}_N) & \cdots & G(\mathbf{a}_L, b_L, \mathbf{x}_N) \end{bmatrix}_{N \times L}.$$

3) Calculate the hidden layer's output weight matrix $\beta_{L \times m}$ by solving the least squares problem:

$$\beta = \mathbf{H}^\dagger \mathbf{T}$$

, where $\mathbf{H}^\dagger$ is the generalized inverse matrix and $\mathbf{T} = \begin{bmatrix} \mathbf{t}_1^T \\ \vdots \\ \mathbf{t}_N^T \end{bmatrix}_{N \times m}$.

---

### E. Constrained Mixed Extreme Learning Machine

The Constrained Mixed Extreme Learning Machine (CMELM) utilizes mixed vectors, containing class-constrained difference vectors and class-constrained sum vectors, to construct the weights from the input layer to the hidden layer. The CMELM firstly generates half numbers of hidden nodes whose weights and biases are constructed with constrained sum vectors, then generates the others whose weights and biases are constructed with constrained difference vectors. The constrained sum vectors are normalized as same as that of CSELM, and the constrained difference vectors are normalized as same as that of CDELM. The normalized sum sample vectors are assigned as the weights from the input layer to the hidden layer. The constrained mixed vectors can be considered as model average (the average model of CDELM and CSELM) because of the linear property of the output layer.

From the above discussion, we can design the CMELM algorithm as Algorithm 5.

**Algorithm 5:** Training of the Constrained Mixed Extreme Learning Machine (CMELM)

**Input:** the training samples $\aleph=\{(\mathbf{x}_i,\mathbf{t}_i)|\mathbf{x}_i \in \mathbf{R}^n, \mathbf{t}_i \in \mathbf{R}^m, i=1,\ldots,N\}$, the hidden node number $L$ and the activation function $G(\mathbf{w},b,\mathbf{x})$.

**Output:** the model parameters of CELM, i.e., the weight matrix $\mathbf{W}_{n \times L}$ and the bias vector $\mathbf{b}_{1 \times L}$ from the input layer to the hidden layer, the weight matrix $\beta_{L \times m}$ from the hidden layer to the output layer.

1) While the number of constrained sum vectors is less than $\lceil L/2 \rceil$

  a) Randomly draw training samples $\mathbf{x}_c'$ and $\mathbf{x}_c''$ from the same class and generate the sum vector $\mathbf{x}_c' + \mathbf{x}_c''$;

  b) Normalize the sum vector by $\mathbf{w} = \dfrac{\mathbf{x}_c' + \mathbf{x}_c''}{\left\|\mathbf{x}_c' + \mathbf{x}_c''\right\|_{L_2}^2}$, and draw the corresponding bias $b$ randomly from [0,1] uniform distribution.

  c) Use the vector $\mathbf{w}$ and bias $b$ to construct the weight matrix $\mathbf{W}_{n \times (\lceil L/2 \rceil)}$ and bias vector $\mathbf{b}_{1 \times (\lceil L/2 \rceil)}$.

2) While the number of chose difference vectors is less than $L - \lceil L/2 \rceil$

  a) Randomly draw training samples $\mathbf{x}_{c1}$ and $\mathbf{x}_{c2}$ from any two different classes respectively and generate the difference vector $\mathbf{x}_{c2} - \mathbf{x}_{c1}$;

  b) Normalize the difference vector by $\mathbf{w} = \dfrac{2(\mathbf{x}_{c2} - \mathbf{x}_{c1})}{\left\|\mathbf{x}_{c2} - \mathbf{x}_{c1}\right\|_{L_2}^2}$, and calculate the corresponding bias $b = \dfrac{(\mathbf{x}_{c1} + \mathbf{x}_{c2})^T (\mathbf{x}_{c1} - \mathbf{x}_{c2})}{\left\|\mathbf{x}_{c2} - \mathbf{x}_{c1}\right\|_{L_2}^2}$.

  c) Use the vector $\mathbf{w}$ and bias $b$ to construct the weight matrix $\mathbf{W}_{n \times (L - \lceil L/2 \rceil)}$ and bias vector $\mathbf{b}_{1 \times (L - \lceil L/2 \rceil)}$.

2) Concatenate the above $\mathbf{W}_{n \times (\lceil L/2 \rceil)}$ and $\mathbf{W}_{n \times (L - \lceil L/2 \rceil)}$ to form the hidden layer nodes weights, and concatenate the $\mathbf{b}_{1 \times (\lceil L/2 \rceil)}$ and $\mathbf{b}_{1 \times (L - \lceil L/2 \rceil)}$ to form the hidden layer nodes biases. Calculate the hidden layer output matrix $\mathbf{H}$ as

$$\mathbf{H} = \begin{bmatrix} \mathbf{h}(\mathbf{x}_1) \\ \vdots \\ \mathbf{h}(\mathbf{x}_N) \end{bmatrix} = \begin{bmatrix} G(\mathbf{a}_1,b_1,\mathbf{x}_1) & \cdots & G(\mathbf{a}_L,b_L,\mathbf{x}_1) \\ \vdots & \cdots & \vdots \\ G(\mathbf{a}_1,b_1,\mathbf{x}_N) & \cdots & G(\mathbf{a}_L,b_L,\mathbf{x}_N) \end{bmatrix}_{N \times L}.$$

3) Calculate the hidden layer's output weight matrix $\beta_{L \times m}$ by solving the least squares problem:

$$\beta = \mathbf{H}^\dagger \mathbf{T}$$

, where $\mathbf{H}^\dagger$ is the generalized inverse matrix and $\mathbf{T} = \begin{bmatrix} \mathbf{t}_1^T \\ \vdots \\ \mathbf{t}_N^T \end{bmatrix}_{N \times m}$.

## IV. PERFORMANCE EVALUATION

In this section, we evaluate the proposed CELMs and compare it with some classifiers, such as ELM, SVM and some related deep learning methods, on both synthetic and real-world datasets. Ten rounds of experiments are conducted for each data set. In each experiment, the training set and the test set are randomly generated using the samples from synthetic datasets and UCI database [39]. The samples from UCI database are normalized to be of zero mean and unit variance. The performances are recorded with the means and the standard deviations of classification accuracies.

In these experiments, we also compare the CELMs with the orthogonal ELM [40], which makes weight vectors orthogonal to each other and biases orthogonal to each other. The aim of this comparison is to compare CELMs with other ELM related methods appeared in literature sufficiently. The code of ELM used in the experiments was downloaded from [14]. In the following figures, the red solid performance curve stands for ELM performance, the green solid curve stands for orthogonal ELM performance, the blue solid curve stands for CDELM performance, the blue dashed curve with triangle markers stands for SELM performance, the brilliant blue solid curve stands for CSELM performance, the brilliant blue dashed curve with triangle markers stands for RSELM performance and the black solid curve stands for CMELM performance. The software used in the experiments is MATLAB R2010a with the Microsoft Windows Serve 2003 operation system. The configuration of hardware is Intel(R) Xeon(R) CPU E5440 @2.83GHz. The total RAM of the server is 32.0 GB, but the experiments cannot take up too much due to other users' usage.

### A. Experiments on Synthetic Dataset

We first evaluate our CELM algorithms on the synthetic dataset of the spiral data. It is illustrated in Fig. 3. To retain the symmetrical shape of the spiral, we normalize the samples into the range [-1, 1] as same as that in [14]. The total number of such generated spiral data is 5000. Two thirds of data samples are used as training set and the rest are used as test set. These data samples are randomly drawn from original spiral data samples. The two sets are randomly generated in each one of the total ten rounds of experiments.

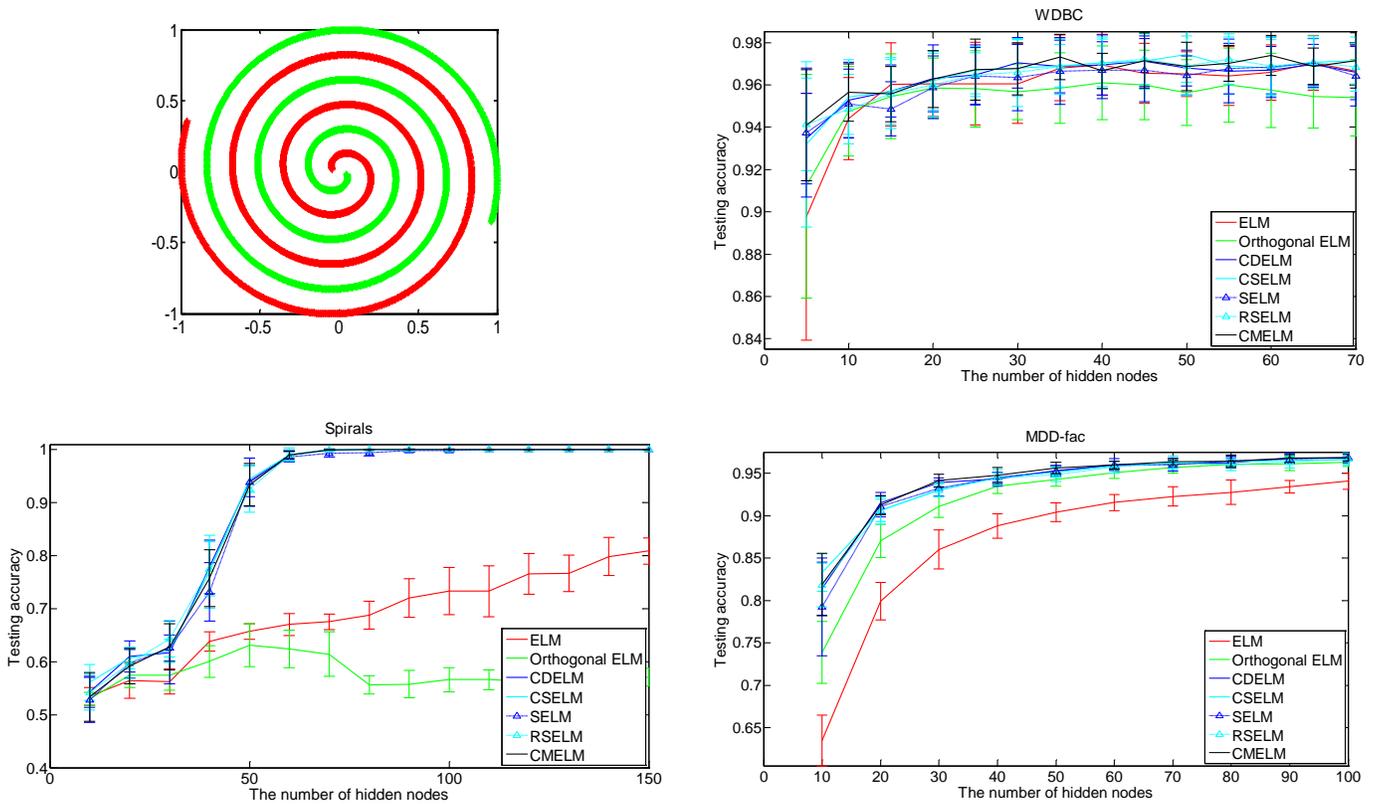

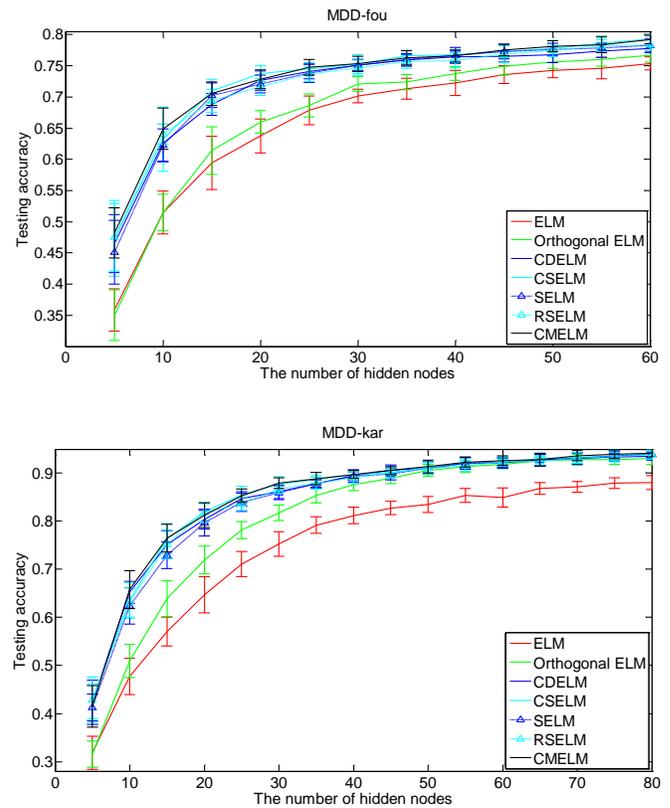

Fig. 3. The spiral synthetic dataset is illustrated at the top side. The performance comparisons are shown at the bottom side.

We compare the performances of CELMs with ELM and orthogonal ELM. The number of hidden nodes is selected from 10 to 150 at a step 10. The performances of these models are illustrated in Fig. 2.

As shown in Fig. 2, CELMs have a perfect performance when the number of hidden nodes reaches a slight larger than 50, while orthogonal ELM drops when the number of hidden nodes is larger than 50. The drop of orthogonal ELM is probably because the orthogonalization in two vectors of 50 dimensions is of no help, even degrades the information in the random weights. The test accuracy of ELM only reaches 0.8 even when the number of hidden nodes is 150. The test accuracies of CELMs are above those of orthogonal ELM and ELM all the time. The result shows that CELMs have the better generalization abilities than orthogonal ELM and ELM. Besides, the variances of these CELMs are becoming quite small with the increase of hidden nodes number and there is little difference among the performances of these CELMs. Therefore, the difference vectors and sample vectors really work from the performance comparison between CELMs and orthogonal ELM and ELM.

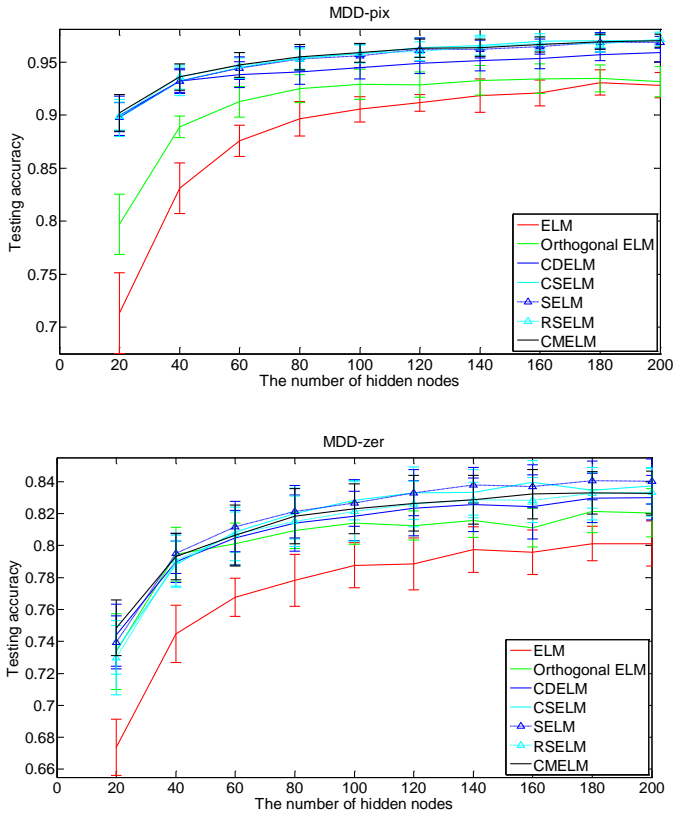

Fig. 4. Experimental results of ELM, orthogonal ELM and CELMs on the six UCI datasets.

### B. Experiments on UCI Datasets

Six datasets from UCI database [39], including Wisconsin diagnostic breast cancer dataset (WDBC) and digit datasets in five different features (MDD-fac, fou, kar, pix, zer), are used for evaluating the proposed CELMs. CELMs are compared with ELM, orthogonal ELM.

In the first experiment, the number of training samples is 2/3 size of the total samples, and the rest is used as test samples. The training and test sets are randomly generated. The comparison of ELM, CELMs and orthogonal ELM in the first experiment is illustrated in Fig. 3. The performances of these models are displayed sufficiently from the trends of the curves in the figure. It can be seen that the test accuracy curves of CELMs are above those of other two methods in the experiment and so the generalization ability of the proposed CELMs is better than other two models on these real world datasets. The samples' distribution prior we introduced makes the efficient use of hidden nodes in CELMs and really helps the classification tasks.

In the second experiment, a benchmark performance evaluation is conducted on the six datasets. The size of the training set is also 2/3 of the total samples and the remaining data is used as the test set. We compare our methods with the regularized ELM, regularized orthogonal ELM, BP network and linear SVM methods. It should be mentioned that the optimization objectives of ELM, orthogonal ELM and CELMs used in the benchmark evaluation are all added a regularized term. Three-fold cross validation is used to select the best regularization factor in $log_{10}$ space from -8 to 8 with the step 1. The number of hidden nodes used in ELM, orthogonal ELM, CELMs and BP network is from 5 to 200 with the step 5 in five MDD data sets. And in WDBC data set, the number of hidden nodes used is from 5 to 100 with the step 5. The selection of hidden nodes number is based on the Fig. 3. Although it is unfair for those methods that are not converged completely, the aim of the proposed methods is to improve the efficiency of hidden neurons. Due to no kernels used in ELM, the SVM used here is linear SVM. The cost factor used in SVM is selected as same as the regularization factor in ELM. The performances of ten rounds of experiments are recorded. The mean of test accuracies and training times are recorded in Table I. The best test accuracies are represented in bold face.

The training method of BP neural network is RPROP [41] due to time and memory problems in the experiments. The liner SVM code used is the MATLAB code obtained from [43].

It can be seen from the Table I that the performances of most CELMs outperform those of other methods. The CELMs improve the performance significantly, whilst retaining the extremely high learning speed of ELM.

### C. Experiments on Large Scale Datasets

We also evaluate the CELMs and ELM on two large size datasets, i.e., MNIST [42] and CIFAR-10 [32]. The MNIST database of handwritten digits contains a training set of 60,000 samples, and a test set of 10,000 samples. It consists of binary images of ten classes and the size of these digits is $28\times28$ pixels. The samples are normalized before input to the CELMs and ELM. The CIFAR-10 dataset contains 60,000 color images in 10 classes, with 6000 image per class. The training set and the test set consist of 50,000 images and 10,000 images respectively. In CIFAR-10 dataset, the standard evaluation pipeline defined in [43] is adopted. First, extract dense $6\times6$ local patches with ZCA whitening and the stride is 1. Second, use threshold coding with $\alpha=0.25$ to encode. The codebook is trained with OMP-1 [44] and the codebook size is 50 in the experiment. Third, average-pool the features on a $2\times2$ grid to form the global image representation.

The performances of CELMs, ELM and orthogonal ELM are illustrated in Fig. 4. In this experiment, these methods are implemented without regularized terms for the sake of evaluating the efficiency of hidden neurons sufficiently.

From Fig. 4, it can be learned the performances of CELMs are about 8 percentages higher than those of ELM averagely on the two datasets. The efficiency of the sample vectors based weighting adopted by SELM can be evaluated as follows. When the same numbers of vectors are used, the method with higher test accuracy means more efficient. In Fig. 4, the performances of CELMs are higher than that of orthogonal ELM when they are not converged. The gaps between the performances of CELMs and orthogonal ELM show that the sample vector based weighting is effective. The curve of CELMs' performance is always above those of orthogonal ELM and ELM in all these datasets, which shows that sample vectors based weighting really helps the efficient use of hidden nodes.

| DATASETS | | WDBC | MDD-fac | MDD-fou | MDD-kar | MDD-pix | MDD-zer |
|---|---|---|---|---|---|---|---|
| ELM | Test accuracy | 0.974 | 0.958 | 0.805 | 0.923 | 0.934 | 0.809 |
| | Training time(s) | 0.01 | 0.03 | 0.03 | 0.03 | 0.03 | 0.04 |
| Orthogonal ELM | Test accuracy | 0.969 | 0.976 | 0.807 | 0.937 | 0.942 | 0.827 |
| | Training time(s) | 0.01 | 0.07 | 0.04 | 0.05 | 0.08 | 0.04 |
| Constrained Difference ELM | Test accuracy | 0.973 | 0.975 | 0.839 | 0.953 | 0.961 | 0.833 |
| | Training time(s) | 0.02 | 0.05 | 0.04 | 0.04 | 0.05 | 0.04 |
| Constrained Sum ELM | Test accuracy | **0.975** | 0.977 | **0.84** | **0.964** | 0.973 | 0.838 |
| | Training time(s) | 0.02 | 0.07 | 0.07 | 0.08 | 0.08 | 0.07 |
| Sample ELM | Test accuracy | 0.973 | **0.978** | **0.84** | **0.964** | 0.970 | **0.841** |
| | Training time(s) | 0.02 | 0.04 | 0.03 | 0.03 | 0.04 | 0.03 |
| Random Sum ELM | Test accuracy | 0.974 | 0.977 | **0.84** | 0.963 | 0.970 | 0.836 |
| | Training time(s) | 0.01 | 0.04 | 0.03 | 0.03 | 0.04 | 0.03 |
| Constrained Mixed ELM | Test accuracy | 0.974 | 0.977 | **0.84** | 0.963 | **0.971** | 0.836 |
| | Training time(s) | 0.03 | 0.07 | 0.07 | 0.07 | 0.08 | 0.06 |
| BP | Test accuracy | 0.972 | 0.616 | 0.428 | 0.416 | 0.525 | 0.378 |
| | Training time(s) | 0.97 | 3.69 | 3.80 | 4.27 | 5.33 | 4.98 |
| Linear SVM | Test accuracy | 0.628 | 0.965 | 0.827 | 0.950 | 0.965 | 0.831 |
| | Training time(s) | 0.73 | 12.37 | 0.43 | 0.151 | 2.88 | 1.41 |

Experiments are conducted on MNIST data set to compare CELMs with ELM, orthogonal ELM and Multi-Layer Extreme Learning Machine (ML-ELM) [40]. The ML-ELM is a Stacked Denoising Auto Encoder (SDAE) model [46] based on ELM. The random feature mapping is used as encoder in ELM and the linear system is used as decoder. The original ML-ELM with three layers of hidden nodes 700-700-15000 can outperform other deep learning methods, such as Deep Belief Network (DBN) [48], Deep Boltzmann Machine (DBM) [48], Stacked Auto Encoder (SAE) and SDAE [46]. In our experiment, the network structure of ML-ELM is set to 700-700-1000 due to the limit of our machine capacity. And the number of hidden nodes used in ELM, orthogonal ELM and CELMs is set to 1000 and 2000 for evaluation. In the experiments, regularization term is used for ELM, orthogonal ELM and CELMs. The parameter selection is the same as before. The average performance is recorded in Table Ⅱ.

From Table Ⅱ, it can be learned when the number of CELMs' hidden nodes is 1000, the test accuracies of CELMs are comparable to ML-ELM, but CELMs have much high learning speed. When the number of CELMs' hidden nodes is 2000, the CELMs outperform the ML-ELM of our implemented version. But the learning speed of CELMs is slower due to the computation complexity of matrix inversion. From the significant performance gap between ELM and CELMs and the similar learning speed of ELM and CELMs, we can conclude the sample based features used in CELMs really help the improvement, whilst retaining fast learning property of

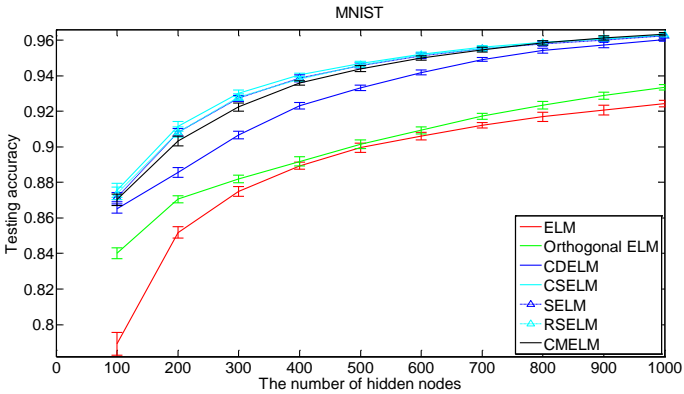

(a)

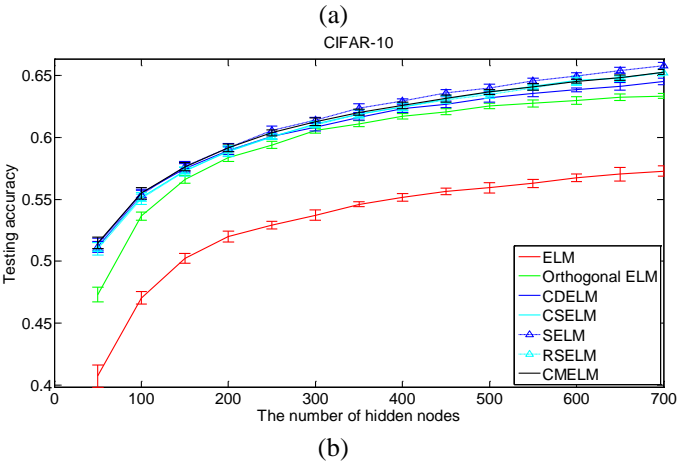

(b)

Fig. 5. Experimental results of ELM, orthogonal ELM and CELMs on the subsets of MNIST and CIFAR-10 datasets.

ELM.

The CELMs are also compared with SVM related methods, e.g., Linear SVM and R$^2$SVM [30], and deep ELM related method, e.g., DCN [31], on CIFAR-10 dataset. The R$^2$SVM is a deep learning model and its building block is a linear SVM model. The outputs of the previous layers are transformed by a random matrix, and then the transformed outputs are added to the original features. The modified features are input into the next layer after transformed by a sigmoid function. The DCN is also a deep learning model, but its building block is an ELM based model, in which parts of the hidden nodes are built with random projection and the other part of hidden nodes are built with RBM weights [31, 45]. Instead of the way that adds the output of previous layer as a bias to the next layer, the outputs of the previous layers are concatenated with the original features sequentially and are input into the next layer in DCN model.

In this experiment, all the 50,000 training samples are used to train the model, and all the 10,000 test samples are used to evaluate the performance. Table III shows the performances of ELM, orthogonal ELM, CELMs, linear-SVM, R$^2$SVM [30] and DCN [31] methods. Note the performances of linear SVM, R$^2$SVM and DCN are cited from [30]. And the experimental conditions of ELM, orthogonal ELM and CELMs, such as the features and the number of used training and test sets, are the same with [30].

TABLE I
AVERAGE CLASSIFICATION ACCURACIES ON UCI DATASETS

TABLE II
PERFORMANCE ON MNIST DATASET

| Algorithms | No. of hidden nodes | Test accuracy | Training Time(s) |
|---|---|---|---|
| ELM | 1000 nodes | 0.930 | 35.39 |
| Orthogonal ELM | 1000 nodes | 0.934 | 37.73 |
| Constrained Difference ELM | 1000 nodes | 0.963 | 34.46 |
| Constrained Sum ELM | 1000 nodes | 0.964 | 35.47 |
| Sample ELM | 1000 nodes | 0.964 | 35.32 |
| Random Sum ELM | 1000 nodes | 0.963 | 35.94 |
| Constrained Mixed ELM | 1000 nodes | 0.964 | 36.46 |
| ELM | 2000 nodes | 0.947 | 107.25 |
| Orthogonal ELM | 2000 nodes | 0.957 | 110.99 |
| Constrained Difference ELM | 2000 nodes | 0.974 | 107.51 |
| Constrained Sum ELM | 2000 nodes | 0.973 | 107.16 |
| Sample ELM | 2000 nodes | 0.974 | 106.91 |
| Random Sum ELM | 2000 nodes | 0.973 | 108.16 |
| Constrained Mixed ELM | 2000 nodes | **0.976** | 106.20 |
| Multi-Layer ELM (ML-ELM) | 2400 nodes (700-700-1000) | 0.968 | 58.23 |

TABLE III
PERFORMANCE ON CIFAR-10 DATASET

| Algorithms | No. of hidden layers /nodes | Test accuracy |
|---|---|---|
| Linear SVM | - | 0.647 |
| R$^2$SVM | 60 layers | 0.693 |
| DCN | Tens to hundreds layers | 0.672 |
| ELM | 7000 nodes | 0.645 |
| Orthogonal ELM | 7000 nodes | 0.683 |
| Constrained Difference ELM | 7000 nodes | 0.720 |
| Constrained Sum ELM | 7000 nodes | 0.728 |
| Sample ELM | 7000 nodes | **0.733** |
| Random Sum ELM | 7000 nodes | 0.727 |
| Constrained Mixed ELM | 7000 nodes | 0.723 |

From the Table III, the CELMs can be found to have the better performances than that of linear-SVM, DCN and R$^2$SVM. The CELMs have the test accuracies of at least 8 percentages higher than that of linear SVM. In [30], the R$^2$SVM has 60 layers, and each layer is a linear SVM after random projection and sigmoid transformation. Although R$^2$SVM and DCN have many layers, the CELMs of one hidden layer have the test accuracy of at least 3 percentages higher than that of these discriminative deep learning methods. Besides, the CELMs can train a model at a case of 50,000 training data well, which suggests that the proposed CELMs can tackle large scale data effectively.

To further understand the feature mapping of ELM, orthogonal ELM and the CELMs, the last experiment is conducted for the visualization of the hidden layer's feature mapping of these methods on MNIST data set. The visualization method, t-SNE [49], is used. The t-SNE is a very ideal visualization tool due to preserving local structure and overcoming the "crowding" problem of mapped data. The s-SNE has a much better visualization effect than other methods, such as PCA [38], LLE [50] and Auto Encoder [47], on MNIST data set. In the experiment, the whole 60,000 training data are used as training set and 2000 data are randomly selected from 10,000 test data to be used as the test set in t-SNE. The number of hidden nodes in ELM, orthogonal ELM and CELMs is 10,000. The visualization is illustrated in Fig. 5.

Seen from the figure, our CELMs can retain the data structure well with the pre-assigned iteration number. The hidden layer's feature mapping in CELMs can separate the data of different classes well. However, after the transformation of 10,000 hidden neurons in ELM and orthogonal ELM, the input data do not converge after many iterations in t-SNE visualization. The phenomenon probably reveals the essence that the constrained random mapping hidden neurons may outperform the completely random hidden neurons in traditional ELMs. It also provides new insights into ELM related research that fast generation of meaningful hidden neurons could boost ELM's performance greatly.

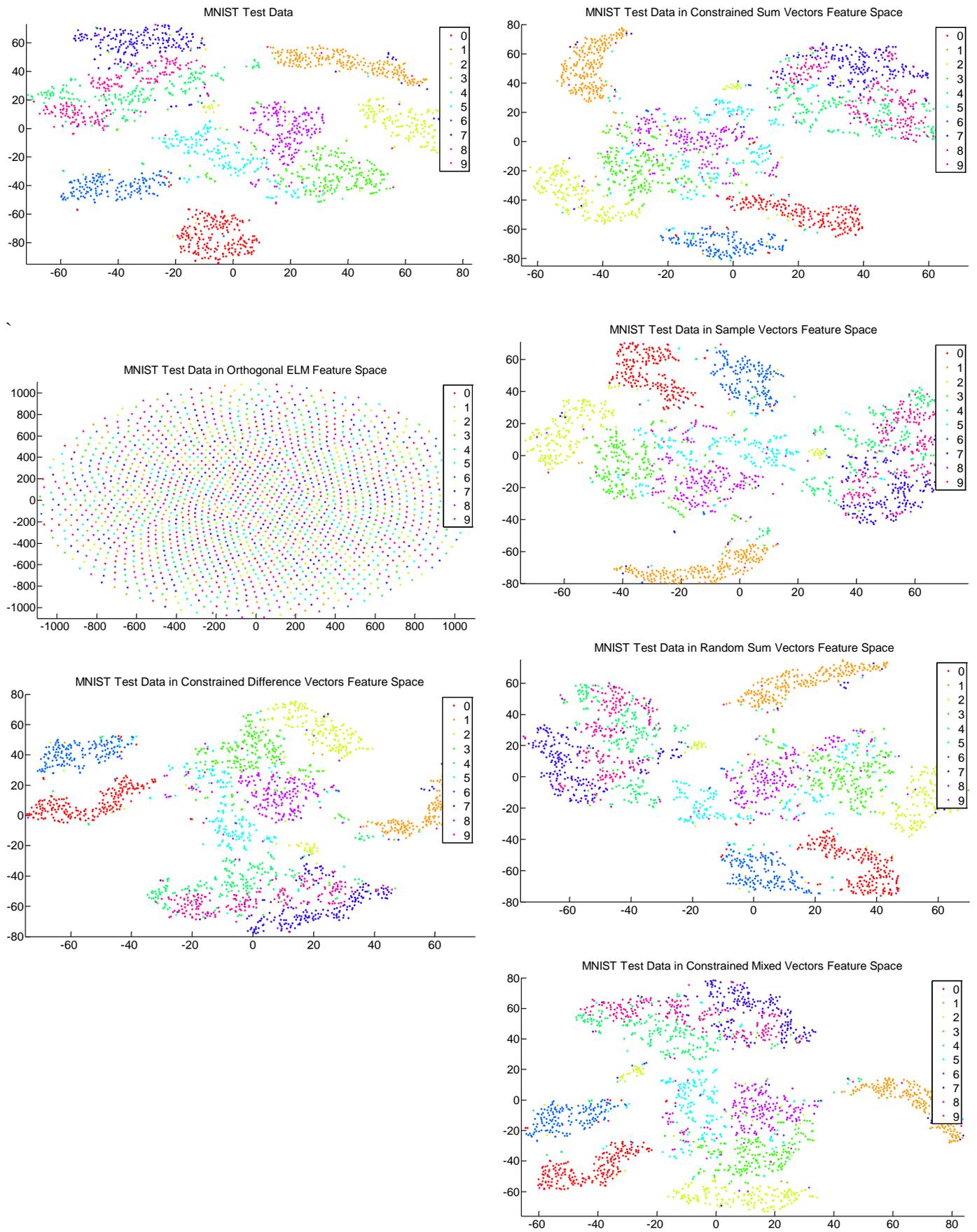

Fig. 6. ELM, orthogonal ELM and the CELMs hidden layer feature visualization on MNIST dataset.

## V. Discussion

We have compared CELMs with several ELM based methods, such as normalized ELM [37], orthogonal ELM, and ELM. The main contributions of the CELMs to the ELM study are, 1) we introduce a constrained hidden weights based on sample distributions, 2) we normalized the hidden weights by the square of their $l_2$ norms, other than $l_2$ norm (The normalized ELM is taken the strategy.). Several work have validated the effectiveness of CELMs [53].

From the experiments of CELMs, we observe that when the number of hidden nodes is small, the CELMs are outperform normalized ELM, orthogonal ELM and ELM greatly. However, when the number of hidden nodes is large, the margins between the ELM and normalized ELM are not that big. From the observation, the constrained weights and normalized strategy both work for the success of CELMs.

On one hand, when the number of hidden nodes is small, the hidden layer is equivalent to dimension reduction. Thus the directions of hidden weights are quite import since they represent the directions to be retained after the reduction. The weights based on sample distribution work well for the dimension reduction [52].

On the other hand, the normalization of hidden weights is important due to the property of sigmoid activation function. The effective response area of sigmoid function is near the zero. When the weights are not normalized and the dimension of inputs is high, the absolute values of many elements in $WX+b$ are very big. Thus the normalization is quite import. And we strongly recommend the operator will be added in the research of ELM in the future.

## VI. Conclusion

To address the inefficient use of hidden nodes in ELM, this paper proposed the novel learning models, CELMs. The CELMs constrain its random weights' generation from a smaller space compared to that of the ELM, i.e., replacing the completely random weight vectors with ones that are randomly drawn from the set of simple linear combination of sample vectors. The main contribution of CELMs is that it introduces sample distribution prior into the construction of the hidden layer to make a better feature mapping and benefit the next layer's linear classification. The effective feature mapping greatly contributes the efficient use of hidden nodes in ELM. Extensive comparisons between CELMs and some related methods on both synthetic and real-world datasets showed that CELMs have better performances in almost all the cases.

However, the CELMs still have some problems that typical ELM owned. One is that CELMs face the over fitting problem when the number of hidden nodes is very large, although CELMs improve the effective use of discriminative hidden nodes. To our relief, the methods in [34, 35, 51] can tackle the problem effectively. Another problem is that the solving of the weights from the hidden layer to the output layer is time-consuming when the number of hidden nodes is very large. This case is much common in large scale applications. Some gradient based solving methods for linear system can tackle the problem iteratively.

The further research will include the study of invariant feature generating for the improvement of CELMs and the experimental verification on CELMs' application to regression problems. The analyses on what kinds of problems that the CELMs will work with and such related theories are also expected to be studied in the future.

## Acknowledgment

The authors would like to thank Dr. L. L. C. Kasun, Dr. H. Zhou, Prof. G.-B. Huang from Nanyang Technological University, Singapore and Prof. C. M. Vong from University of Macau, Macau for their kindly help with Multi-Layer Extreme Learning Machine (ML-ELM).